\documentclass[sort&compress, numafflabel]{elsarticle}

\usepackage[]{natbib}
\usepackage[breaklinks,hidelinks]{hyperref}
\usepackage{times}
\usepackage{latexsym}

\usepackage{geometry}
\usepackage{subfiles}
\usepackage{multirow}
\usepackage{array}
\usepackage{hyphenat}
\usepackage{subcaption}
\usepackage{longtable}
\usepackage{graphicx}
\usepackage{adjustbox}
\usepackage{enumitem}
\usepackage{url}

\usepackage[color=yellow, textsize=small]{todonotes}
\usepackage{bm}
\usepackage{booktabs}
\usepackage{float}
\usepackage[english]{babel}
\usepackage{blindtext}
\usepackage{textcomp}
\usepackage{soul,color}
\usepackage{pifont}

\makeatletter
\def\ps@pprintTitle{%
 \let\@oddhead\@empty
 \let\@evenhead\@empty
 \def\@oddfoot{}%
 \let\@evenfoot\@oddfoot}
\makeatother
\usepackage{epstopdf}
\AppendGraphicsExtensions{.tif}

\usepackage{titlesec}
\titleformat{\section}
      {\normalfont\bfseries}
      {\thesection}
      {0ex}
      {\MakeUppercase}
\titleformat{\subsection}
      {\normalfont\bfseries}
      {\thesection}
      {0ex}
      {}
\titleformat{\subsubsection}
      {\normalfont}
      {\thesection}
      {0ex}
      {}

\newcommand\blfootnote[1]{%
  \begingroup
  \renewcommand\thefootnote{}\footnote{#1}%
  \addtocounter{footnote}{-1}%
  \endgroup
}

\usepackage{setspace}
\newif\ifsubfile
\subfiletrue
\newif\iftif
\tiffalse

\usepackage[utf8]{inputenc}
\usepackage[T1]{fontenc}
\usepackage{lineno}
\usepackage[numbers]{natbib}
\usepackage{subfiles}
\usepackage{graphicx}
\usepackage{float}
\usepackage{hyperref}
\usepackage{mathtools}
\usepackage{changepage}
\usepackage{caption}
\usepackage{multirow}
\usepackage{titlesec}
\usepackage{amsmath}

\usepackage{makecell}

\usepackage[math]{cellspace}
\usepackage{tikz}

\usepackage{url}

\usepackage{colortbl}
\linenumbers
\nolinenumbers

\usepackage{listings}
\usepackage{xcolor}

\title{Generalizable and Scalable Multistage Biomedical Concept Normalization Leveraging Large Language Models}

\author[jhu]{Nicholas J Dobbins\textsuperscript{+}\blfootnote{\textsuperscript{+}Corresponding author: Nicholas Dobbins, PhD, MLIS, Biomedical Informatics and Data Science, Department of Medicine, Johns Hopkins University, 2024 East Monument St. S 1-200, Baltimore, Maryland 21205 USA; nic.dobbins@jhu.edu}}

\ead{}

\address[jhu]{Biomedical Informatics and Data Science, Department of Medicine, Johns Hopkins University, Baltimore, MD, USA}

\begin{document}
\subfilefalse

\newpageafter{author}

\begin{abstract}

\noindent\textbf{Background:} Biomedical entity normalization is critical to biomedical research because the richness of free-text clinical data, such as progress notes, can often be fully leveraged only after translating words and phrases into structured and coded representations suitable for analysis. Large Language Models (LLMs), in turn, have shown great potential and high performance in a variety of natural language processing (NLP) tasks, but their application for normalization remains understudied.  \\

\noindent\textbf{Methods:} We applied both proprietary and open-source LLMs in combination with several rule-based normalization systems commonly used in biomedical research. We used a two-step LLM integration approach, (1) using an LLM to generate alternative phrasings of a source utterance, and (2) to prune candidate UMLS concepts, using a variety of prompting methods. We measure results by $F_{\beta}$, where we favor recall over precision, and F1. \\

\noindent\textbf{Results:} We evaluated a total of 5,523 concept terms and text contexts from a publicly available dataset of human-annotated biomedical abstracts. Incorporating GPT-3.5-turbo increased overall $F_{\beta}$ and F1 in normalization systems +9.5 and +7.3 (MetaMapLite), +13.9 and +10.9 (QuickUMLS), and +10.5 and +10.3 (BM25), while the open-source Vicuna model achieved +10.8 and +12.2 (MetaMapLite), +14.7 and +15 (QuickUMLS), and +15.6 and +18.7 (BM25). \\

\noindent\textbf{Conclusions:} Existing general-purpose LLMs, both propriety and open-source, can be leveraged at scale to greatly improve normalization performance using existing tools, with no fine-tuning.

%                           Vicuna         GPT
%MetaMapLite 43.4   54.8   +12.2 +10.8   +7.3  +9.5
%QuickUMLS   30.3   39.2   +15   +14.7   +10.9 +13.9
%BM25        28.1   46.3   +18.7 +15.6   +10.3 +10.5

\end{abstract}

\begin{keyword}
UMLS, normalization, biomedical concepts, entity linking, large language model
\end{keyword}

\maketitle

\pagebreak

\section*{Introduction}
\label{sec:background}

Biomedical entity normalization, also known as entity linking or grounding, is the processing of mapping spans of text, such as conditions, procedures, or medications, into coded representations, such as Unified Medical Language System (UMLS) codes. Normalization is critical to biomedical research because the richness of free-text data, such as information within progress notes, can often be fully leveraged only after mapping words and phrases into structured, coded representations suitable for analysis. For example, identifying patients with Type 2 Diabetes Mellitus using free-text narratives could be challenging with keyword search alone, given the variety of possible phrasings ("T2DM", "Hyperglycemia", "Glucose intolerance", etc.) Searching instead for a normalized representation (e.g., UMLS "C0011860") simplifies this process while greatly improving recall. Coded terms can subsequently be linked to related terms within ontologies and so on. Recent state-of-the-art Large Language Models (LLMs), in turn, have shown great potential and high performance in a variety of natural language processing (NLP) tasks, but their application for normalization remains understudied. Moreover, while biomedical informaticians and researchers often leverage rule-based systems, such as MetaMapLite \cite{demner2017metamap} or cTAKES \cite{savova2010mayo} for normalization, few studies have evaluated the use of LLMs working \textit{in concert} with commonly used existing normalization systems.

In this study, we evaluate the use of two widely use LLMs, one closed-source (GPT-3.5-turbo) \cite{chatgpt} and one open (Vicuna-13b \cite{vicuna2023}, a fine-tuned variation of Llama \cite{touvron2023llama}), alongside widely used normalization techniques and libraries within the informatics community, on a large human-annotated corpus of biomedical abstracts and UMLS concepts \cite{mohan2019medmentions}. We aim to contribute to improving performance for a common scenario within biomedical informatics research: the need to extract normalized concepts from a large corpus of documents, where fine-tuning a domain-specific model for the task is not practical or possible (for example due to the lack of a gold standard annotation or time). In such cases, many researchers use applications such as MetaMapLite, QuickUMLS \cite{soldaini2016quickumls}, cTAKES \cite{savova2010mayo}, CLAMP \cite{soysal2018clamp}, which tend to be rule-based and relatively fast and scalable in concept extraction, but often with only moderate recall and relatively low precision. At a basic level, we seek to explore the question: \textit{How can comparatively small, widely available LLMs be leveraged to improve upon baseline biomedical concept normalization performance?} We examine a variety of prompting strategies with a focus on two specific areas where LLMs may aid improvement alongside existing normalization systems. All code used in this study is available at \url{https://github.com/ndobb/llm-normalization/}\footnote{Git repository will be made publicly available upon article acceptance}.

\section*{Background and Significance}
\label{sec:background}

The process of concept normalization from biomedical free-text documents has been studied extensively, and a number of widely used normalization systems exist. Such systems tend to be rule-based, parsing text and matching based on lexical and syntactic heuristics \cite{aronson2010overview, demner2017metamap, soldaini2016quickumls, savova2010mayo}, though more recent systems, such as the CLAMP toolkit \cite{soysal2018clamp}, also incorporate machine learning based models for sentence boundary detection and named entity recognition.

Related to normalization, Narayan \textit{et al} \cite {narayan2022can} explored the use of LLMs using OpenAI's GPT-3 \cite{mann2020language} on publicly available datasets to evaluate entity matching between datasets (e.g., determining if two products are the same based on their descriptions), error detection, and data imputation. Peeters \textit{et al} \cite{peeters2023using, peeters2023entity} similarly expanded this line of research using GPT-3.5 and systematically explored various prompting strategies. Other work in this area has been driven by ontology researchers aiming to determine how LLMs may be leveraged for entity alignment and relation prediction (e.g,. \textit{is-a}, \textit{subsumes}) to automate ontology generation and error checking \cite{hertling2023olala, arora2023linktransformer, matentzoglu2023mappergpt, babaei2023llms4ol}. 

Within the health domain, Yang \textit{et al} \cite{yang2023integrating} used a Retrieval Augmented Generation (RAG) based approach \cite{lewis2020retrieval} to inject UMLS-derived context within prompts to improve question-answering performance. Specific to normalization, Soroush \textit{et al} \cite{soroush2024large} analyzed the use of GPT models for generating ICD-9, ICD-10, and CPT codes from text-descriptions, finding both GPT-3.5-turbo and GPT-4 to perform relatively poorly, with accuracy under 50\%. Soroush \textit{et al's} findings suggests that LLMs \textit{alone} may be inappropriate tools for concept normalization. This challenge may be even moreso for UMLS concepts (as opposed to ICD-10), which include a far larger number of longer, total codes which themselves carry no inherent meaning and imply no relation to other codes (for example, the sequence of characters for the UMLS code "C0011860" are essentially arbitrary, unlike codes which begin with "E" in ICD-10, which relate in some fashion to diabetes mellitus). 

\subsection*{Key Contributions}
As current SoTA LLMs alone perform poorly at normalization, we seek to explore \textbf{how well they can improve baseline normalization performance when used alongside and augmenting widely-used existing normalization software}. To do so, we:

\begin{enumerate}
    \item \textbf{Establish baseline precision, recall, F1 and $F_{\beta}$} performance using several common normalization applications on a large set of human-annotated, normalized condition mentions.
    \item Use both open- and closed-source LLMs for \textbf{synonym and alternative phrasing generation}, which we further normalize in order to maximize recall. 
    \item Explore prompting strategies for \textbf{normalized concept pruning} to subsequently balance precision.
    \item Utilize combinations of normalization systems and LLMs to \textbf{robustly evaluate end-to-end normalization strategies aimed at maximizing $F_{\beta}$}, a harmonic mean of precision and recall weighted towards recall.
\end{enumerate}

We designed our experiments with a focus on \textbf{practicality} and \textbf{scale}, intended to inform biomedical research efforts requiring normalization of large-scale unannotated clinical text repositories while minimizing potential cost. Our results can be leveraged by researchers with various goals in mind, such maximizing recall or precision, or in understanding cost and performance strategies in various prompting strategies.

\section*{Materials and Methods}
\label{sec:methods}

\subsection*{Language Models}

Biomedical concept normalization efforts often involve thousands or even millions of documents \cite{xu2010comprehensive, wu2012unified}. In such projects the need for normalization processes at speed and without significant cost (e.g., less than 1 second and under \$0.01 per document) are thus necessary. We chose two language models, one publicly available and one proprietary, as we believed they achieved a reasonable tradeoff between strong (though not SoTA \cite{li2024comparing, meyer2024comparison, rizzo2024performance, sun2024large}) performance, inference speed, and cost. 

\begin{enumerate}
    \item \textbf{GPT-3.5-turbo} is the primary model behind ChatGPT \cite{chatgpt} at the time of this writing. GPT-3.5-turbo has been demonstrated to perform well on a variety of tasks within the clinical domain \cite{liu2023utility, rao2023assessing}, including authoring letters to patients \cite{ali2023using}, decision support \cite{liu2023using}, medical question-answering \cite{johnson2023assessing}, interpreting radiology reports \cite{jeblick2023chatgpt} and various clinical NLP tasks \cite{li2024chatgpt, wagner2023accuracy, hu2024improving}. We used the \textit{gpt-3.5-turbo-0125} model within OpenAI for our experiments.
    \item \textbf{Vicuna} is an open-source model fine-tuned from the original LlaMA model \cite{touvron2023llama} using user-shared responses from ChatGTP \cite{vicuna2023}. As Vicuna was thus tuned to respond similarly to ChatGPT but is freely available and also comparatively smaller, we chose Vicuna as a reasonable alternative capable of running locally without the need for significant infrastructure. We used the quantized \textit{vicuna-13b-v1.5.Q4\_K\_M.gguf} model\footnote{https://huggingface.co/TheBloke/vicuna-13B-v1.5-16K-GGUF} for our experiments.
\end{enumerate}

\subsection*{Normalization Systems}

We chose three methods and software libraries as our baseline normalization systems:

\begin{enumerate}
    \item \textbf{MetaMapLite} \cite{demner2017metamap} is a Java-based implementation of the original MetaMap algorithm \cite{aronson2010overview}. MetaMapLite uses a Lucene-based \cite{bialecki2012apache} dictionary lookup approach for normalizing concepts indexed by UMLS concept source text, abbreviations, and source term types.
    \item \textbf{QuickUMLS} \cite{soldaini2016quickumls} is a lightweight Python-based implementation of the \textit{CPMerge} algorithm \cite{okazaki2010simple}. QuickUMLS is designed to achieve similar recall and precision to MetaMap and cTAKES but much faster.
    \item \textbf{BM25}, or more formally \textbf{Okapi BM25}, is a widely used ranking and retrieval algorithm \cite{whissell2011improving}. We used a Python implementation of BM25 from the \textit{retriv} library\footnote{https://github.com/AmenRa/retriv}. We indexed BM25 using the preferred terms of UMLS concepts related to diseases and conditions.
\end{enumerate}

%'Congenital Abnormality', 'Acquired Abnormality', 'Injury or Poisoning', 'Pathologic Function', 'Disease or Syndrome',  'Mental or Behavioral Dysfunction', 'Cell or Molecular Dysfunction', 'Experimental Model of Disease', 'Anatomical Abnormality',  'Neoplastic Process', 'Sign or Symptom'

\subsection*{Dataset}

MedMentions \cite{mohan2019medmentions} is a human-annotated dataset of over 4,000 biomedical abstracts. Each abstract contains mentions of biomedical concept terms and their most appropriate corresponding UMLS concept ID. We chose MedMentions because of both the high-quality of the annotations and the large relative size of the corpus.

As the focus of this study is to inform low-resource scenarios where researchers are unable to train or fine-tune models for normalization, we utilized only the test set of MedMentions for evaluation purposes. The test set consists of 839 abstracts with 70,405 UMLS concepts. As we aimed to evaluate our LLM-augmented normalization methods rather than exhaustively validate the corpus using the entirety of the UMLS, we limited our experiments to 5,523 UMLS concept annotations related to disease and conditions.

\subsection*{Normalization Strategy}

We aimed to leverage the LLMs described to improve the baseline normalization performance of MetaMapLite, QuickUMLS, and BM25 in a relatively simple pipeline-like process. Importantly, we assume a scenario where named entities within a given corpus are already known, but not normalized. In other words, we imagine a case where a named entity recognition (NER) algorithm has already identified candidate token spans within a corpus (for example, "cystic fibrosis" at character indices 250-265 in document 1) but UMLS concepts are not known. We seek to augment traditional normalization processes with the following strategies:

\begin{enumerate}
    \item \textbf{Synonym and Alternate Phrasing Generation} - We prompt an LLM to generate alternative phrasings or synonyms of a given input text span to be normalized. We then use our normalization systems to attempt to normalize both the source text span, as well as candidate alternative spans. This step seeks to maximize recall.
    \item \textbf{Candidate Concept Pruning} - After normalization, we again prompt an LLM to filter out inappropriate concepts, using both the preferred term and semantic type of a given candidate concept as well as surrounding text context of the original identified span. This step seeks to improve precision.
\end{enumerate}

\begin{figure}[H]
  \includegraphics[scale=0.78]{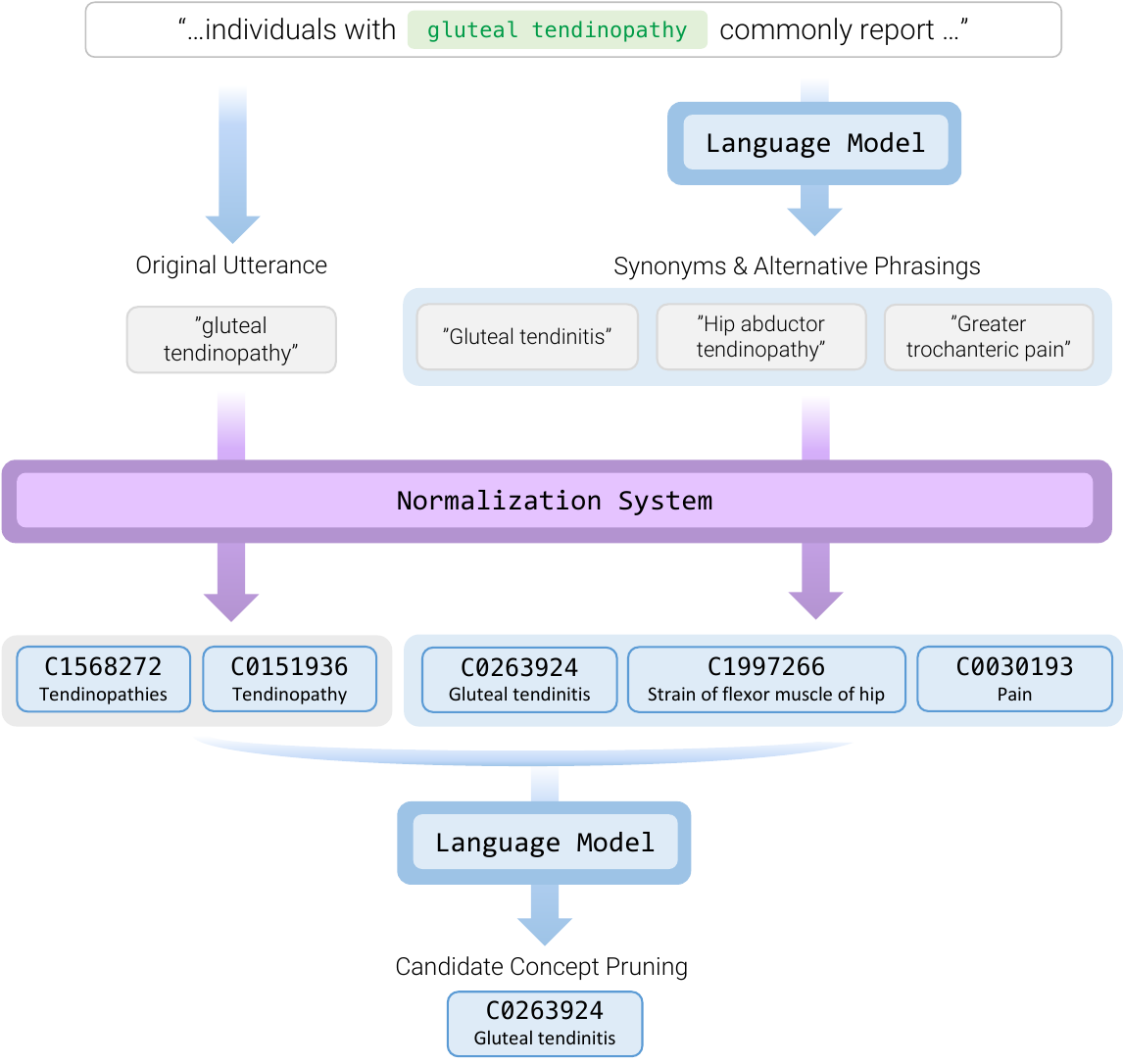}  
\caption{Diagram of our multi-stage normalization strategy.}
\label{fig_workflow}
\end{figure}

Figure \ref{fig_workflow} shows a visual example of this process.

\subsection*{Evaluation}

Our evaluation process was as follow:

\begin{enumerate}
    \item Using all three normalization systems, we established baseline normalization results by individually executing each on all condition concepts in the test set of the MedMentions corpus.
    \item We prompted both LLMs to generate candidate alternative phrasings and synonyms for all test set concepts, then used our normalization systems to subsequently normalize the additional concepts as well, focusing on recall.
    \item Next, we sought to identify an optimal prompting strategy for concept pruning. To do so, we experimented using a random sample of 1,000 test set concepts and the Vicuna model with various prompt structures. These were:
    \begin{enumerate}
        \item \textbf{Multiple Choice} - We included all candidate concepts in a single prompt, instructing the LLM to output a list of appropriate concepts. Within the multiple choice approach, we further experimented with outputting UMLS concept \textbf{concept IDs (CUIs) versus indices} in which they were presented (e.g., "C0010674" vs. "2"). We hypothesized that simpler index-based output may may perform slightly better than CUIs. We additionally evaluated \textbf{chain-of-thought} prompting, instructing the model to "think step-by-step" and provide reasoning for a given output.
        \item \textbf{Binary Choice} - We evaluated each candidate concept independently, with an additional LLM prompt and response for each concept. We hypothesized that a simpler prompt and question may also lead to better performance, though at the cost of far more overall prompts needed.
    \end{enumerate}

    Figure \ref{fig_prompt_strategies} shows examples of these strategies.

    \begin{figure}[H]
      \centering
      \includegraphics[scale=0.53]{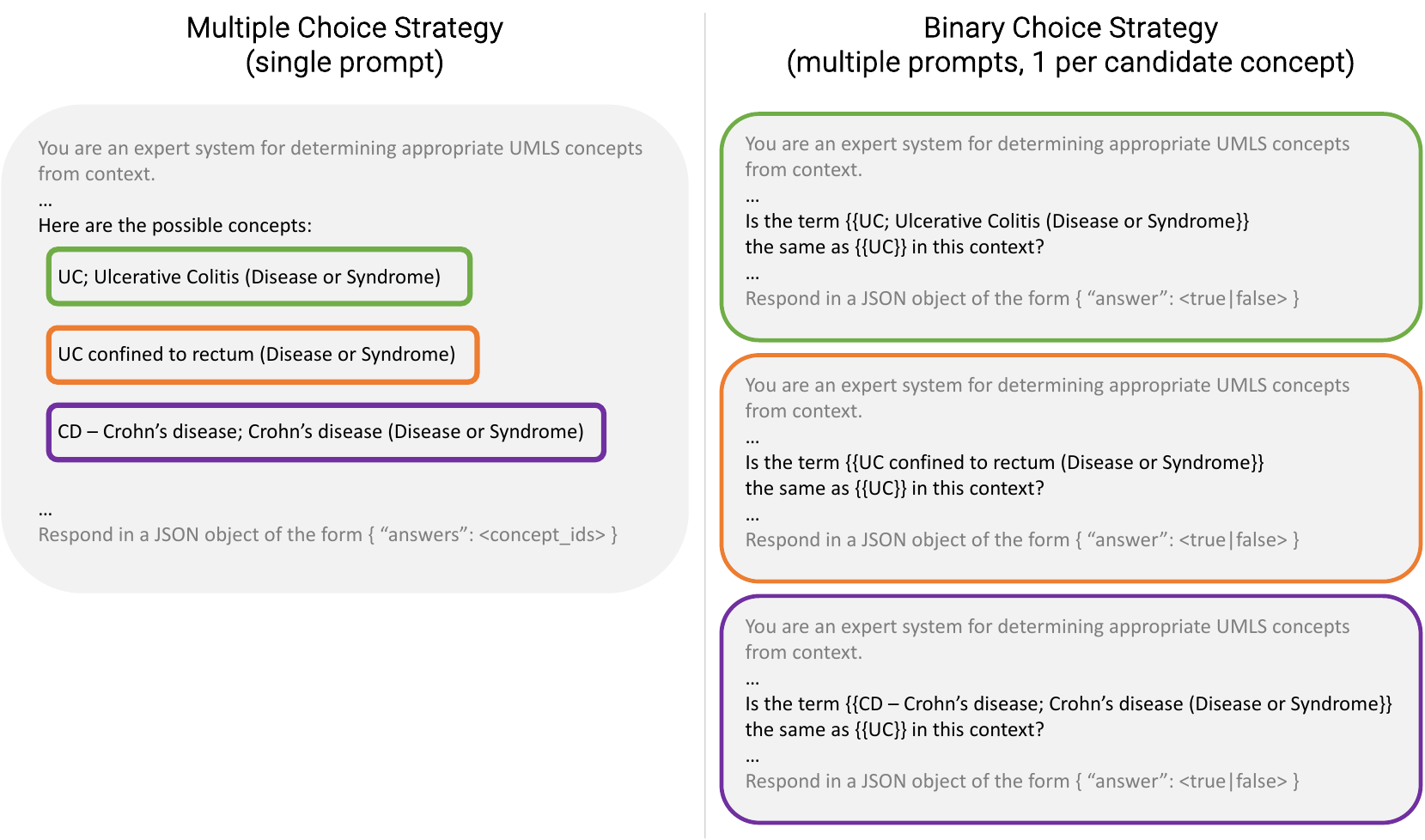}  
    \caption{Visual example of our Multiple Choice and Binary Choice prompting strategies.}
    \label{fig_prompt_strategies}
    \end{figure}
    
    In addition, for each prompt strategy combination, we also experimented with a post-processing step to \textbf{always accept the first candidate concept}, even if rejected in the concept pruning phase. We refer to this strategy hereafter as \textbf{Top1}.

    For scoring purposes, we make the explicit assumption that recall is ultimately more important than precision in this task. For example, many downstream analyses of normalized, extracted data often look for specific CUI values (e.g., identifying patients with heart failure "C0018801" as clinical trial candidates). Thus false positive CUIs may be ignored in such cases with relatively low harm. We therefore consider $F_{\beta}$ as our metric to optimize, with recall having greater weight than precision ($\beta = 2$):
    
    \[ F_{\beta} = (1 + \beta^2) * \frac{precision*recall}{(\beta^2*precision)+recall} \]
    
    For context we also provide the traditional F1 harmonic mean of precision and recall:

    \[ F1 = 2*\frac{precision*recall}{precision+recall} \]
    
\end{enumerate}

\section*{Results}
\label{sec:results}

Table \ref{tbl_expr_1_2} shows results of our baseline and synonym and alternate phrasing phrasing experiments. As our results at this stage include candidate concepts from our LLM-generated alternate phrasings without pruning, the improvements shown are thus the upper bound of recall. We prompted each LLM to return at most 3 alternate phrasings. For each alternate phrasing, we accepted only the first normalized concept returned, if any. While both LLMs showed recall improvement over baseline results with each normalization system, GPT-3.5-turbo showed a greater recall improvement for each over Vicuna. With the exception of BM25 (which already had a reasonably high recall of 77.7\% at baseline), each normalization system showed an improvement in recall of over 10\% for both LLMs. 

\begin{table}[h!]
    \small
    \centering
    \def\arraystretch{1.4}
\begin{tabular}{c|cccc|cccc|cccc}
                           & \multicolumn{4}{c|}{Baseline} 
                           & \multicolumn{4}{c|}{Vicuna} 
                           & \multicolumn{4}{c}{GPT-3.5-turbo} \\
     \textbf{Norm. System} & \textbf{P} & \textbf{R} & \textbf{F1} & {$\mathbf{F_\beta}$}
                           & \textbf{P} & \textbf{R} & \textbf{F1} & {$\mathbf{F_\beta}$}
                           & \textbf{P} & \textbf{R} & \textbf{F1} & {$\mathbf{F_\beta}$} \\
     \hline
     MetaMapLite  & 32.2 & 66.5 & 43.4 & 54.8    & 6.2  & 76.6 & 11.5 & 23.4    & 10.8 & 78.7 & 19.0 & 34.8 \\
     QuickUMLS    & 21.9 & 49.0 & 30.3 & 39.2    & 7.0  & 62.9 & 12.7 & 24.2    & 8.2  & 66.2 & 14.6 & 27.4 \\
     BM25         & 17.7 & 77.7 & 28.1 & 46.3    & 12.2 & 82.3 & 21.4 & 38.2    & 13.0 & 84.8 & 22.5 & 40.2 \\
\end{tabular}

    \caption{Results of our first experiments using LLMs to generate and normalize synonyms and alternative phrasings of an initial utterance. We aimed to improve recall over baseline results.}
    \label{tbl_expr_1_2}
\end{table} 

Next, we aimed to determine optimal prompt and scoring strategies for concept pruning, using variants of our prompt strategies. Given the large number of potential LLM responses needed due to various combinations of prompt strategies, we chose to use only the Vicuna LLM and a random subset of 1,000 examples from the test set. Results are shown in Table \ref{tbl_expr_3}.

\begin{table}[h!]
    \small
    \centering
    \def\arraystretch{1.4}
\begin{tabular}{ll|cccc|cccc}
        & & \multicolumn{8}{c}{\textbf{Vicuna}} \\
        & & \multicolumn{4}{c|}{\textbf{LLM-chosen only}} & \multicolumn{4}{c}{\textbf{LLM-chosen+Top1}} \\
     \textbf{Norm. System} & \textbf{Prompt Strategy} 
        & \textbf{P} & \textbf{R} & \textbf{F1} & {$\mathbf{F_\beta}$}
        & \textbf{P} & \textbf{R} & \textbf{F1} & {$\mathbf{F_\beta}$} \\
     \toprule

    \multirow{6}{*}{MetaMapLite}   
        & Multiple-Choice+Index+CoT & 45.7 & 65.4 & 53.8 & 60.2 & 46.1 & 69.7 & 55.5 & 63.2 \\
        & Multiple-Choice+Index & 43.9 & 63.9 & 52.0 & 58.6 & 44.8 & 70.4 & 54.8 & 63.2 \\
        & Multiple-Choice+CUI+CoT & 42.6 & 70.6 & 53.1 & 62.4 & 41.6 & 72.6 & 52.9 & 63.2 \\
        & Multiple-Choice+CUI & 39.3 & 69.6 & 50.2 & 60.3 & 39.0 & 72.3 & 50.7 & 61.8 \\
        & Binary-Choice+CoT & 44.7 & 70.2 & 54.6 & 63.0 & 41.9 & 74.2 & 53.6 & 64.3 \\
        & Binary-Choice & 46.8 & 67.1 & 55.1 & 61.7 & 44.3 & 74.0 & 55.4 & 65.3 \\
    \hline
     
    \multirow{6}{*}{QuickUMLS}   
        & Multiple-Choice+Index+CoT & 39.9 & 54.7 & 46.1 & 50.9 & 40.4 & 57.2 & 47.4 & 52.8 \\
        & Multiple-Choice+Index & 39.7 & 56.2 & 46.5 & 51.9 & 39.7 & 58.0 & 47.1 & 53.1 \\
        & Multiple-Choice+CUI+CoT & 40.5 & 57.8 & 47.6 & 53.3 & 40.1 & 58.7 & 47.6 & 53.7 \\
        & Multiple-Choice+CUI & 37.5 & 57.7 & 45.5 & 52.1 & 37.3 & 59.0 & 45.7 & 52.9 \\
        & Binary-Choice+CoT & 38.6 & 59.3 & 46.8 & 53.6 & 36.0 & 63.7 & 46.0 & 55.2 \\
        & Binary-Choice & 39.1 & 56.6 & 46.2 & 51.9 & 37.0 & 63.5 & 46.8 & 55.5 \\
    
    \hline
    \multirow{6}{*}{BM25}       
        & Multiple-Choice+Index+CoT & 36.8 & 52.1 & 43.1 & 48.1 & 37.1 & 53.5 & 43.8 & 49.2 \\
        & Multiple-Choice+Index & 36.7 & 52.1 & 43.1 & 48.1 & 37.1 & 53.9 & 43.9 & 49.4 \\
        & Multiple-Choice+CUI+CoT & 40.4 & 54.4 & 46.4 & 50.9 & 40.4 & 55.2 & 46.7 & 51.4 \\
        & Multiple-Choice+CUI & 39.5 & 54.8 & 45.9 & 50.9 & 39.2 & 55.3 & 45.9 & 51.1 \\
        & Binary-Choice+CoT & 33.7 & 76.2 & 46.7 & 60.9 & 31.4 & 80.1 & 45.1 & 61.1 \\
        & Binary-Choice & 35.0 & 73.3 & 47.4 & 60.1 & 32.8 & 79.0 & 46.4 & 61.6 \\
\end{tabular}

    \caption{Results of experiments to determine an optimal prompting strategy for concept pruning. We used a randomly-chosen subset of 1,000 test set concepts and contexts.}
    \label{tbl_expr_3}
\end{table} 

For each normalization system, as the binary prompt strategy with chain-of-thought showed the best performance by $F_{\beta}$ among our test set sample using LLM-chosen UMLS concepts, we chose to use this in our final experiment. As automatically including the first candidate concept also improved recall (though at the expense of some precision), we also included the "Top1" strategy in our final experiment. Results are shown in Table \ref{tbl_expr_4}. 

Somewhat surprisingly, with higher precision than GPT-3.5-turbo for each normalization system, the smaller Vicuna model achieved the highest F1 scores over GPT-3.5-turbo in all experiments. $F_{\beta}$ scores were generally closer with the exception of BM25, where Vicuna achieved notably better $F_{\beta}$ than GPT-3.5-turbo (CoT: +3.9 and CoT+Top1: +5.1). Compared to baseline normalization $F_{\beta}$ results in our first experiment (shown in Table \ref{tbl_expr_1_2}), the addition of the Vicuna model for end-to-end alternative phrasing and best pruning strategy led to over 10\% improvement in all normalization systems (MetaMapLite: +10.8, QuickUMLS: +14.7, BM25: +15.6).

\begin{table}[h!]
    \small
    \centering
    \def\arraystretch{1.4}
\begin{tabular}{ll|cccc|cccc}
                           & & \multicolumn{4}{c|}{Vicuna} 
                           & \multicolumn{4}{c}{GPT-3.5-turbo} \\
     \textbf{Norm. System} & \textbf{Strategy}
                           & \textbf{P} & \textbf{R} & \textbf{F1} & {$\mathbf{F_\beta}$}
                           & \textbf{P} & \textbf{R} & \textbf{F1} & {$\mathbf{F_\beta}$} \\
     \hline
     \multirow{2}{*}{MetaMapLite} & Binary-Choice+CoT
                                           & 46.4 & 68.3 & 55.3 & 62.4
                                           & 38.1 & 75.6 & 50.7 & 63.1 \\
                                  & Binary-Choice+CoT+Top1  
                                           & 44.3 & 74.6 & 55.6 & 65.6
                                           & 37.5 & 78.3 & 50.7 & 64.3 \\
     \hline
     \multirow{2}{*}{QuickUMLS}   & Binary-Choice+CoT      
                                           & 38.6 & 56.3 & 45.8 & 51.5
                                           & 30.8 & 63.5 & 41.5 & 52.3 \\
                                  & Binary-Choice+CoT+Top1 
                                           & 35.8 & 61.7 & 45.3 & 53.9 
                                           & 30.0 & 65.8 & 41.2 & 53.1 \\
     \hline
     \multirow{2}{*}{BM25}        & Binary-Choice+CoT     
                                           & 35.2 & 73.6 & 47.6 & 60.4
                                           & 25.6 & 81.1 & 38.9 & 56.5 \\
                                  & Binary-Choice+CoT+Top1 
                                           & 33.2 & 79.1 & 46.8 & 61.9
                                           & 24.9 & 83.7 & 38.4 & 56.8 \\
\end{tabular}

    \caption{Results of our final experiment to apply our best-performing prompting strategy using all normalization systems and LLMs.}
    \label{tbl_expr_4}
\end{table} 

Full prompts for all experiments are included in the Appendix.

\section*{Discussion}
\label{sec:discussion}

Our results demonstrate that existing LLMs can be leveraged to greatly increase performance of widely-used biomedical normalization systems without fine-tuning. Our two-step process of leveraging LLMs for alternate phrasing generation and subsequent concept pruning demonstrates improvements to $F_{\beta}$ (best: +15.6) and F1 (best: +19.5) in each combination LLM and normalization system we experimented with, as well as higher recall in all experiments with the exception of one (BM25 using Vicuna with Binary-Choice+CoT prompting strategy). 

Moreover, we demonstrate that these results can be achieved using non-SoTA, smaller models, both open-source and proprietary. Our results show that the publicly-available Vicuna 13b quantized model can achieve results that surpass GPT-3.5-turbo in $F_{\beta}$ and F1 under certain circumstances. As these smaller models can respond relatively quickly and at relatively low cost (the \textit{gpt-3.5-turbo-0125} model costs \$1.50 per 1M tokens at the time of this writing), they also lend themselves to scalability, in concert with normalization systems.

\subsection*{Limitations}

This study had a number of limitations. First, we used only a single dataset, MedMentions, and only concepts related to diseases and conditions. It is possible that other datasets, textual contexts, domains and concept semantic types (e.g., medications) may show different results. As the MedMentions test set includes over 5,000 concepts from a large variety of biomedical abstracts, however, we argue that the dataset is nonetheless useful and reasonable for establishing the efficacy of our methods. Next, we evaluated only two large language models, and potentially there are a growing list of others that may be included. As we intended to demonstrate that existing, readily-available, smaller models encode sufficient biomedical knowledge while also remaining inexpensive and highly responsive, we argue these models are highly suitable to common research situations where a large corpora of documents require normalization at scale while model fine-tuning is not possible or desirable. Additionally, we assume a situation where named entities within a corpus are already identified but not yet normalized, which may not always be the case. As reasonably well-performing NER models, including open-source, are readily-available, we feel this is a reasonable assumption.

\subsection*{Future work}

In future work, we intend to apply these approaches to existing corpora of other domains (such as progress notes), as well as evaluate cost and scaling factors in greater detail.

\section*{Conclusion}
\label{sec:conclusion}

This study demonstrates that smaller, widely-available large language models can be readily leveraged alongside existing normalization software to achieve greater precision, recall, $F_{\beta}$ and F1 in clinical documents. We empirically evaluated two language models in combination with three widely-user normalization systems using a variety of prompting and processing strategies. The methods discussed here can be readily adapted and used by research teams.

%TC:ignore 
\section*{Acknowledgements} 

We thank Meliha Yetisgen, Fei Xia, Ozlem Uzuner, and Kevin Lybarger for their thoughts and suggestions in the planning phase of this project. This study was supported in part by the National Library of Medicine under Award Number R15LM013209 and by the National Center for Advancing Translational Sciences of National Institutes of Health under Award Number UL1TR002319. The content is solely the responsibility of the authors and does not necessarily represent the official views of the National Institutes of Health. 

\section*{Author contributions statement}

NJD conceived of and executed all experiments and wrote the manuscript.

\section*{Competing interests}

NJD is a consultant of TriNetX, LLC.

%\bibliography{main}

\section*{Appendix}
\label{sec:appendix}

\subsection*{Prompt for Synonym and Alternate Phrasing}
\begin{figure}[H]
  \centering
  \includegraphics[scale=0.78]{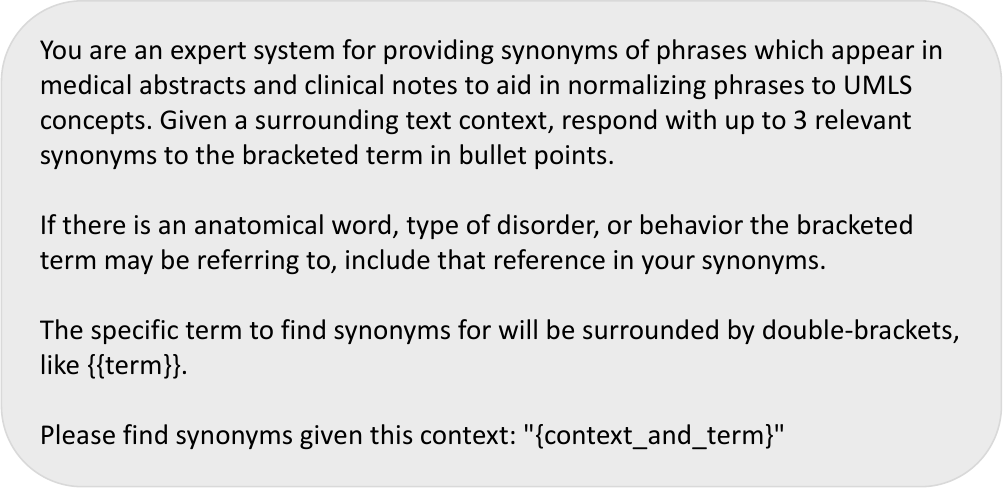}  
\caption{Prompt used for synonym and alternate phrasing generation.}
\label{fig_prompt1}
\end{figure}

\subsection*{Prompt for Multi-Choice Concept Pruning}
\begin{figure}[H]
  \centering
  \includegraphics[scale=0.78]{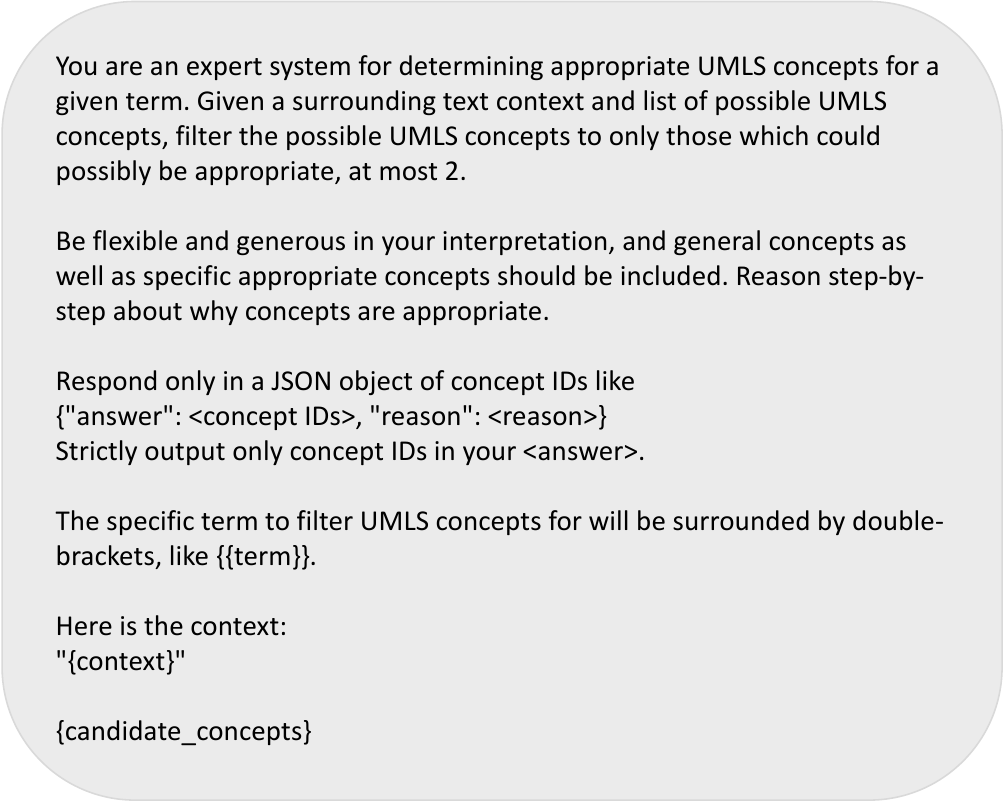}  
\caption{Prompt used for concept candidate pruning. Multi-candidate concepts are presented in a single prompt, and a list of appropriate concepts expected in the response. This example shows the chain-of-though prompt variant.}
\label{fig_prompt2}
\end{figure}

\subsection*{Prompt for Binary Concept Pruning}
\begin{figure}[H]
  \centering
  \includegraphics[scale=0.78]{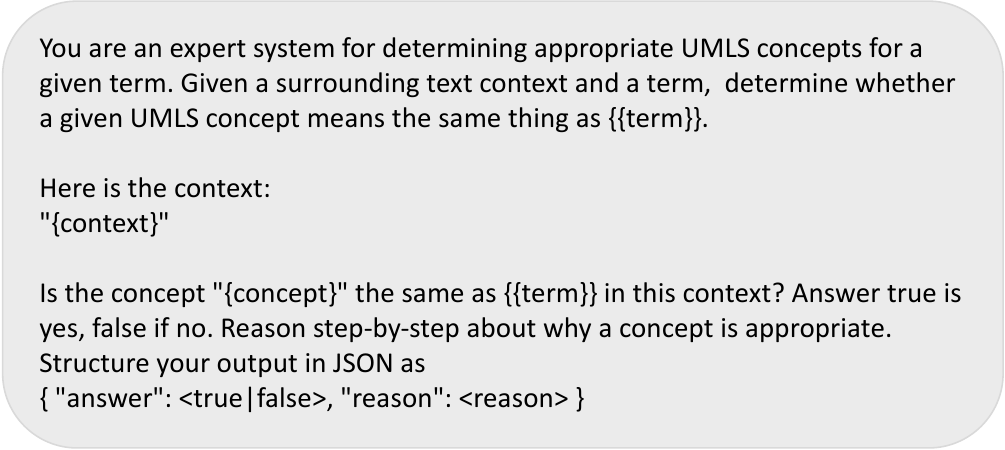}  
\caption{Prompt used for binary concept candidate pruning. A single candidate concept is presented in a single prompt, with a single LLM response needed for all candidate concepts. This example shows the chain-of-though prompt variant.}
\label{fig_prompt3}
\end{figure}

%TC:endignore 

\end{document}